\def\year{2020}\relax
\documentclass[letterpaper]{article} 
\usepackage{aaai20}  
\usepackage{times}  
\usepackage{helvet} 
\usepackage{courier}  
\usepackage[hyphens]{url}  
\usepackage{graphicx} 
\usepackage{graphicx}  
\usepackage{amsmath}
\usepackage{subcaption}

\newcommand{\loss}{\mathcal{L}}
\newcommand{\source}{\mathcal{D}_s}
\newcommand{\target}{\mathcal{D}_t}
\newcommand{\FeatureExtractor}{\mathrm{FE}}
\newcommand{\Discriminator}{\mathrm{D}}
\newcommand{\Prober}{\mathrm{P}}
\newcommand{\Classifier}{\mathrm{C}}
\newcommand{\Group}{\mathrm{G}}
\newcommand{\softmax}{\mathrm{softmax}}

\newcommand{\xvec}{\mathbf{x}}
\newcommand{\yvec}{\mathbf{y}}
\newcommand{\hvec}{\mathbf{h}}
\newcommand{\bvec}{\mathbf{b}}
\newcommand{\svec}{\mathbf{s}}
\newcommand{\dvec}{\mathbf{d}}
\newcommand{\Wmat}{\mathbf{W}}

\DeclareMathOperator*{\argmax}{argmax}

\title{Improving Domain-Adapted Sentiment Classification by\\ Deep Adversarial Mutual Learning}
\author{Qianming Xue\textsuperscript{\rm 1}, Wei Zhang\textsuperscript{\rm 1}\thanks{Wei Zhang is corresponding author.}, Hongyuan Zha\textsuperscript{\rm 2}\\ 
\textsuperscript{\rm 1}School of Computer Science and Technology, KLATASDS-MOE, East China Normal University\\
\textsuperscript{\rm 2}School of Computational Science and Engineering, Georgia Institute of Technology\\
xueqianming200@hotmail.com, zhangwei.thu2011@gmail.com, zha@cc.gatech.edu 
}

\begin{document}

\maketitle

\begin{abstract}
Domain-adapted sentiment classification refers to training on a labeled source domain to well infer document-level sentiment on an unlabeled target domain.
Most existing relevant models involve a feature extractor and a sentiment classifier, where the feature extractor works towards learning domain-invariant features from both domains, and the sentiment classifier is trained only on the source domain to guide the feature extractor.
As such, they lack a mechanism to use sentiment polarity lying in the target domain.
To improve domain-adapted sentiment classification by learning sentiment from the target domain as well, we devise a novel deep adversarial mutual learning approach involving two groups of feature extractors, domain discriminators, sentiment classifiers, and label probers.
The domain discriminators enable the feature extractors to obtain domain-invariant features.
Meanwhile, the label prober in each group explores document sentiment polarity of the target domain through the sentiment prediction generated by the classifier in the peer group, and guides the learning of the feature extractor in its own group.
The proposed approach achieves the mutual learning of the two groups in an end-to-end manner. 
Experiments on multiple public datasets indicate our method obtains the state-of-the-art performance, validating the effectiveness of mutual learning through label probers.
\end{abstract}

\section{Introduction}\label{sec:intro}
Domain-adapted sentiment classification aims at training on a labeled source domain to well infer document-level sentiment on an unlabeled target domain.
It is a natural intersection of researches on sentiment classification~\cite{2012Liu-Book} and unsupervised domain adaptation~\cite{Ben-DavidBCKPV10}, and has attracted much attention with the flourish of online review platforms, such as Amazon, Yelp, etc.
On the one hand, the total amount of reviews is exceedingly huge, which brings opportunities to pursue effective sentiment classification models~\cite{Tang-emnlp15}.
On the other hand, a large number of review domains (e.g., product categories in Amazon) make it intractable to manually annotate enough data in each domain for training domain-specific models.
Thus developing automatically domain-adapted methods is imperative in this area.
It is worth noting that this task is somewhat more challenging than some other cross-domain sentiment classification tasks which require a few labeled data in target domains~\cite{ZhangHPJ18}. 

A straightforward solution to the task is to directly apply the sentiment classification models trained on a source domain to a target domain.
It does not obtain any training guidance from target domains.
However, this is not ideal since it ignores the semantic gap between different domains.
Taking Movie and Food domains as examples, the first domain contains common sentiment words such as ``romantic'', ``violent'', and ``dramatic''.
By contrast, the second includes opinion words like ``tasty'' and ``delicious''.
The potentially small set intersection of domain-dependent sentiment words indicates that there is huge potential for improving the naive solution.
As such, it is important to build a domain-adapted sentiment classification model to perform well.

Existing relevant studies could be attributed into two categories: two-stage approaches~\cite{BlitzerDP07,glorot2011domain,Ziser-naacl28,Ziser-acl19} and end-to-end models~\cite{dann,das,acan}. 
The two-stage approaches typically construct unsupervised feature extractors or manually select pivot features across domains in the first stage.
And in the second stage, a sentiment classifier is trained on the labeled source domain. 
Yet the first stage could not be directly guided by the ground-truth, and the selection of pivot features is a little too empirical and costly. 
On the other hand, benefiting from the advanced learning techniques such as adversarial learning~\cite{GaninL15} and maximum mean discrepancy~\cite{gretton2012kernel}, end-to-end models overcome the above issues by training domain-variant feature extractors and sentiment classifiers holistically, without relying on pivot features.
However, despite the promising results they have made, there still remains a major limitation in most of the end-to-end models: data from target domains is not fully utilized.
That is, their sentiment classifiers and feature extractors overlook the sentiment polarity lying in the review text of target domains. 
One relevant study~\cite{das} in this regard obtains pseudo-labels in target domains by a self-ensemble bootstrapping technique to train its sentiment classifier and feature extractor.
However, the pseudo-labels are asynchronously generated by the earlier version of the sentiment classifier, which is a weaker classifier compared with its current version, and possibly limits the effectiveness of training.

In this paper, we propose a novel learning approach, named deep adversarial mutual learning (DAML).
DAML improves domain-adapted sentiment classification by learning sentiment polarity from an unlabeled target domain.
This is partially inspired by the recently proposed deep mutual learning (DML)~\cite{dml} for supervised single-domain tasks, where two classification models teach each other through their synchronously inferred sentiment label distributions which complement true labels.
The rationality behind mutual learning is that these models can learn collaboratively and transfer their own learned knowledge to each other throughout the training process.
This makes the models robust to the noise in the data.
We extend mutual learning to the unsupervised cross-domain sentiment classification scenario through DAML. 
It involves two groups of feature extractors, domain discriminators, sentiment classifiers, and label probers.
Different from standard mutual learning, we leverage the label prober in each group to learn pseudo sentiment label distributions of documents in the target domain, which are generated by the classifier from the other group.
Meanwhile, the label probers guide the learning of feature extractors in their corresponding groups by gradient back-propagation.
Through the above manner, probers act as bridges between classifiers and extractors, ensuring sentiment information in the target domain to be used by them.
Another advantage of probers is that they free classifiers from aligning with each other, which is required by standard mutual learning and harms performance in a fully unsupervised scenario (see Figure~\ref{fig:ml-tr-acc}).
In addition, we leverage gradient reverse layers (GRL)~\cite{dann} to learn domain-variant feature extractors with domain discriminators.
The two groups can be regarded as two models which are mutually learned in an end-to-end manner to improve generalization on the target domain.
We summarize the main contributions of this work as follows:
\begin{itemize}
\item We address the learning of sentiment polarity in the target domain, which is largely overlooked by previous studies.
We propose DAML which combines the merits of adversarial learning and mutual learning, as well as the introduced label probers which learn from classifiers and guide the learning of extractors. 
To our knowledge, this is the first study of adapting mutual learning for domain-adapted sentiment classification. 

\item We evaluate DAML on multiple datasets with different origins.
The extensive experiments show that DAML achieves the state-of-the-art performance.
We also validate the better results of mutual learning with label probers than directly applying standard mutual learning.
As a byproduct, we will release the source code of our approach\footnote{https://github.com/SleepyBag/DAML}.
\end{itemize}

\section{Related Work}
In this section, we briefly discuss domain adaptation, sentiment classification, and deep mutual learning, to highlight the key differences in this research.

\textbf{Domain adaptation.}
Domain adaptation has been a long-standing attractive research topic due to its real applications where labeled data is only available in a source domain~\cite{pan2009survey}.
It is widely recognized that the distribution gap between source domain and target domain is the fundamental challenge.
To address this, early instance-based approaches reweigh each source example following the idea of importance sampling to match the target data distribution
\cite{huang2007correcting}.
Recent advancements in learning domain-variant feature representations include adversarial learning~\cite{GaninL15} which trains an extractor to fool a domain classifier, and maximum mean discrepancy (MMD)~\cite{GrettonSSSBPF12} which measures the degree of domain shift and minimizes it.

In the specific scenario of domain-adapted sentiment classification, pivot-based methods~\cite{BlitzerDP07,YuJ16,Ziser-naacl28,Ziser-acl19}
first heuristically select domain-shared pivot features and then use them to learn the correspondence of domain-specific sentiment words.
However, such an empirical selection is a little costly and its inaccuracy would be delivered to the following learning step.
Some other studies such as~\cite{glorot2011domain} adopt a similar two-stage procedure by learning an unsupervised domain-variant feature extractor (e.g., stacked denoising auto-encoder~\cite{VincentLBM08}) in the first stage and then train a sentiment classifier by taking the obtained features as input.
Due to the nature that the first stage of the above methods is not guided by sentiment labels, there is a potential to improve them by learning in an end-to-end manner. 
Some recent approaches~\cite{dann,das,acan} employ adversarial learning or MMD to fulfill the end-to-end learning.
However, as discussed previously, they inevitably suffer from the limitations of ignoring sentiment polarity lying in target domains or not leveraging them in an effective manner.
In comparison, our work proposes the novel DAML approach to address the above issue, which is the most important contribution of this paper.

\textbf{Sentiment classification.}
Despite the aforementioned studies about domain-adapted sentiment classification, a large amount of efforts have been devoted to the development of well-performed single-domain sentiment classification models~\cite{ZhangW15}, especially deep learning based ones, including recursive neural networks~\cite{SocherPWCMNP13}, convolutional neural networks~\cite{Kim14}, and recurrent neural networks~\cite{Tang-emnlp15}, etc.
To our knowledge, hierarchical attention networks (HAN)~\cite{hrnn}
is the state-of-the-art sentiment classification model, according to the performance comparison shown in~\cite{Chen-EMNLP2016,Wu-AAAI18},
As such, our work takes the main part of HAN, except the last feed-forward output layer, as our feature extractor. 

\textbf{Deep mutual learning.}
Deep mutual learning~\cite{dml} is a recently developed learning approach in the background of model distillation~\cite{hinton2015distilling}, but it does not use a larger teacher model for guiding the training process.
Apart from learning to fit the true labels, DML makes classifiers simultaneously learn from each other by mimicking others' inferred label distributions and finds its applications~\cite{kanaci2019multi,wu2019mutual}.
However, none of existing studies have investigated its power in domain-adapted classification tasks.
Later we show in the experiments that a naive combination of adversarial learning with standard mutual learning does not improve the performance in the task.

\section{Our Approach}
\begin{figure}[!t]
  \begin{center}
  \includegraphics[width=.46\textwidth]{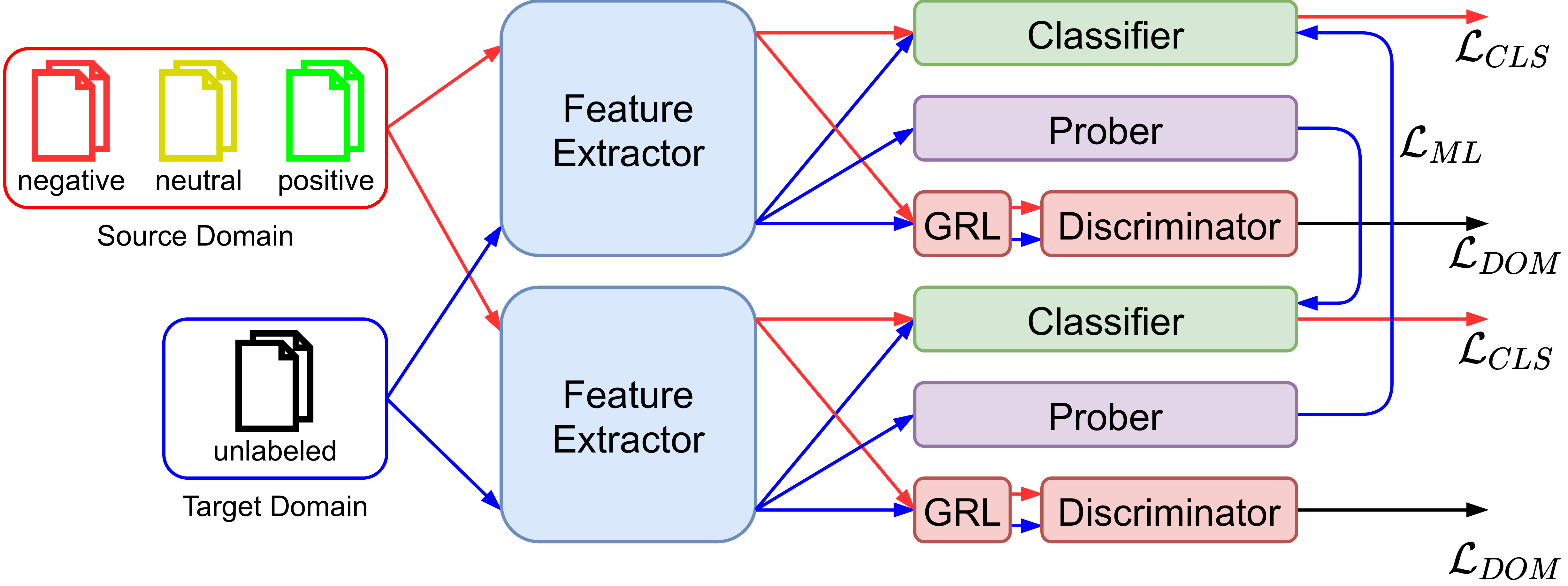}
  \caption{The architecture of deep adversarial mutual learning.
  The red lines and blues lines indicate the information flow from source domain and the target domain, respectively, while the black lines correspond to both domains. 
  $\loss_{DOM}$, $\loss_{ML}$, and $\loss_{CLS}$ denotes domain discriminative loss, mutual learning based loss, and source domain sentiment classification loss, respectively.
}
\label{fig:DAMT}
\end{center}
\end{figure}

In domain-adapted sentiment classification, we are given a labeled source domain $\source=\{(\xvec_i^s,\yvec_i^s)\}_{i=1}^{n^s}$ with $n^s$ documents and an unlabeled target domain $\target=\{\xvec_i^t\}_{i=n^s+1}^{n^s+n^t}$ with $n^t$ documents, where $\xvec$ is the original word sequence representation in a document and $\yvec$ is the corresponding one-hot encoding of sentiment label.
The goal of the task is to learn a well-performed sentiment classification model which is adapted to the target domain by leveraging the examples from $\source$ and $\target$.
To this end, we present a deep adversarial mutual learning approach with the architecture shown in Figure~\ref{fig:DAMT}.

\noindent \textbf{Approach Overview:} The proposed approach consists of two groups of feature extractors, domain discriminators, sentiment classifiers, and label probers. 
The two groups have exactly the same structures but are with their own parameters, wherein each group is associated with both the source and target domains.
The feature extractors take documents from the source domain and the target domain as input, and eventually obtain document representations which are fed into the other components in each group.
Domain discriminators, together with gradient reverse layers, ensure the extractors to get domain-invariant features by the guidance of domain discriminative loss $\loss_{DOM}$.
The prober in each group is associated with the classifier in the other group through the optimization of mutual learning based loss $\loss_{ML}$.
In addition, each classifier also corresponds to a classification loss $\loss_{CLS}$ w.r.t. the labeled source domain.

\subsection{Feature Extraction}
We use HAN as the feature extractors to produce the representations of sentiment documents.
HAN takes a hierarchical structure to mirror the word-level and sentence-level structures of documents.
Assume the $i$-th document from either $\source$ or $\target$ contains $l_i$ sentences and each sentence $j\in\{1,\cdots,l_i\}$ has $m_{i,j}$ words.
HAN first leverages a bidirectional GRU~\cite{BahdanauCB14} to get contextualized word-level representations as follows,
\begin{equation}
\hvec_{i,j,k} = \mathrm{BiGRU}(\xvec_{i,j,k}),~k\in\{1,\cdots,m_{i,j}\}.  
\end{equation}
To capture the importance of each word in constructing a sentence vector $\svec_{i,j}$, word-level attention mechanism is used and defined as follows:
\begin{eqnarray}\label{eq:word-att}
&&\alpha_{i,j,k} = \frac{\exp\big(\omega_w^\top\tanh{(\Wmat_w\hvec_{i,j,k}+\bvec_w)}\big)}{\sum_{k'}\exp\big(\omega_w^\top\tanh{(\Wmat_w\hvec_{i,j,k'}+\bvec_w})\big)}, \nonumber\\
&&\svec_{i,j} = \sum_{k}\alpha_{i,j,k}\hvec_{i,j,k},
\end{eqnarray}
where $\alpha_{i,j,k}$ quantifies the word importance weight. 

Afterwards, another bidirectional GRU is used to model $\svec_{i,j}$  to get contextualized sentence representations $\hvec_{i,j}$ $( j\in\{1,\cdots,l_{i}\})$.
Given this, the document-level representation $\dvec_{i}$ is obtained through sentence-level attention mechanism in a similar fashion as Eq.~\ref{eq:word-att}, i.e.,  $\dvec_{i} = \sum_{j}\alpha_{i,j}\hvec_{i,j}$.

To summarize the procedure of feature extraction, we denote it as $\dvec_{i} = \FeatureExtractor(\xvec_i;\Theta_{FE})$ where $\Theta_{FE}$ covers all the parameters involved in the above computations.

\subsection{Sentiment Classification}
We define a sentiment classifier $\Classifier(\dvec_i; \Theta_C)$ which contains one to several layers of perceptrons and is given as follows:
\begin{equation}
\Classifier(\dvec_i; \Theta_C) = \softmax(\Wmat_c\dvec_i + \bvec_c),
\end{equation}
where $\Wmat_c$ and $\bvec_c$ are learnable parameters and the dimension of output equals to the number of sentiment labels. 
Given this, the classification loss on the labeled source domain is based on cross-entropy, given as:
\begin{equation}
    \loss_{CLS} = \sum_{i=1}^{n^s}\yvec_i^s\log(\Classifier(\dvec_i)).
\end{equation}

\subsection{Domain Adaptation}
To empower the feature extractors with the ability of learning domain-invariant representations, we consider the adversarial learning method.
In particular, we introduce a domain discriminator $\Discriminator$, which is trying to figure out from which domain a document vector comes.
On the contrary, the feature extractor $\FeatureExtractor$ aims to fool $\Discriminator$.

We define D as a multi-layer perceptron with parameters $\Theta_{D}$, which regards the previously obtained document vector as input and outputs a scalar indicating the probability of a document being from the source domain.
For document $\xvec_i$, we set $z_i=1$ if it belongs to the source domain, and $z_i=0$ for the target domain. 
Based on this, a min-max game is played to optimize the parameters $\Theta_{FE}$ and $\Theta_{D}$ as follows:
\begin{equation}
\Theta_{FE}, \Theta_{D} = \argmax_{\Theta_{FE}} \min_{\Theta_{D}} \loss_{DOM}
\end{equation}
where $\loss_{DOM}= -\sum_{i=1}^{n^s+n^t}\big(z_i\log\Discriminator(\dvec_{i})+(1-z_i)\log(1-\Discriminator(\dvec_{i}))\big)$.

As suggested in \cite{dann}, this min-max game is implemented by a gradient
reversal layer, which reverses
$\frac{\partial \loss_{DOM}}{\Theta_{FE}}$ into
$-\eta \frac{\partial\loss_{DOM}}{\Theta_{FE}}$ during the gradient back-propagation process. 
$\eta$ is a controllable hyper-parameter.
Once $\FeatureExtractor$ is able to prevent the discriminator from distinguishing data from the source or target domain, it is supposed to be successfully trained to retain only domain-independent information to some extent.

\subsection{Integration with Mutual Learning}
So far, we are able to utilize the target domain from the aspect of modeling its data distribution.
However, it is still far from satisfactory since the involved sentiment information is not explicitly captured. 
A natural solution is to generate pseudo-labels for the target domain to guide the learning of the sentiment classification model. 

Motivated by mutual learning~\cite{dml} in supervised single-domain tasks, we first consider a direct way of incorporating standard mutual learning (sML) into our framework.
We extend the aforementioned feature extractor, sentiment classifier, and domain discriminator to form two groups (models), i.e., $\Group_1 = (\FeatureExtractor_1, \Classifier_1, \Discriminator_1; \Theta_{G_1})$ and $\Group_2 = (\FeatureExtractor_2, \Classifier_2, \Discriminator_2; \Theta_{G_2})$, where $\Theta_{G_1}$ and $\Theta_{G_2}$ cover all the parameters of the groups, i.e., $\Theta_{G_1} = \{\Theta_{FE_1}, \Theta_{C_1}, \Theta_{D_1}\}$ and $\Theta_{G_2} = \{\Theta_{FE_2}, \Theta_{C_2}, \Theta_{D_2}\}$.
Correspondingly, they have $\loss_{CLS_1}$ and $\loss_{DOM_1}$, $\loss_{CLS_2}$ and $\loss_{DOM_2}$, respectively. 
In consequence, we define the loss of standard mutual learning in our problem setting as follows:
\begin{eqnarray}
  \loss_{sML_1} &=& \sum_{i=n^s+1}^{n^s+n^t} \mathrm{D}_{KL}\big(\Classifier_2(\dvec_i) \| \Classifier_1(\dvec_i)\big), \label{eq:sml_1}\\ 
  \loss_{sML_2} &=& \sum_{i=n^s+1}^{n^s+n^t} \mathrm{D}_{KL}\big(\Classifier_1(\dvec_i) \| \Classifier_2(\dvec_i)\big),\label{eq:sml_2}
\end{eqnarray}
where $\mathrm{D}_{KL}$ is Kullback Leibler (KL) divergence.
Moreover, the two groups have their own hybrid objectives which are defined as follows:
\begin{eqnarray}
  \loss_{G1} &=&  \loss_{CLS_1} + \lambda_{D}\loss_{DOM_1} + \lambda_{M}  \loss_{sML_1}, \label{eq:sml_loss1}\\ 
  \loss_{G2} &=&  \loss_{CLS_2} + \lambda_{D}\loss_{DOM_2} + \lambda_{M}  \loss_{sML_2}, \label{eq:sml_loss2}
\end{eqnarray}
where $\lambda_{D}$ and $\lambda_{M}$ are parameters to control the relative influence of the losses in the whole objectives. 
An alternating optimization strategy is used for mutually learning the two groups.
That is, in each mini-batch, $\frac {\partial \loss_{G_1}} {\partial \Theta_{G_1}}$ and $\frac {\partial \loss_{G_2}} {\partial \Theta_{G_2}}$ are computed and propagated back to update model parameters.

It is intuitive that due to different parameter initialization, the feature extractor and sentiment classifier in each group try to obtain some different abilities (e.g., focusing on some special parts of latent feature spaces) to pursue similar results of its peer group, by the guidance of mutual learning.
As such, they can transfer some exclusive knowledge to the peers.
It is worth noting that Eq.~\ref{eq:sml_1} and~\ref{eq:sml_2} only guide each group with the other's predicted labels, but without telling them how to obtain these pseudo-labels, thus not forcing them to obtain very similar feature representations.
However, the way in which these pseudo-labels are applied to the unsupervised domain-adaptation scenario is somewhat questionable.
This is because the requirement that the two classifiers are made to trust the pseudo-labels generated by each other (see Eq.~\ref{eq:sml_1} and~\ref{eq:sml_2}) is a little too restricted in domain adaptation, where the two classifiers actually cannot see any labeled data in the target domain.
As a result, the classifiers might be not strong enough to teach each other, which turns out that they are misled by the counterparts' predictions and cannot achieve improved classification performance (see Figure~\ref{fig:test-acc} for validation).

To avoid to degrade the classification performance and meanwhile retain the ability of learning sentiment polarity from the target domain, we introduce another component called label prober ($\Prober$) to each group, which has the same architecture as the sentiment classifier but comes with its own learnable parameters.
The goal of label probers is not for sentiment classification like sentiment classifiers.
Instead, each prober probes sentiment information learned by the other group in the target domain, and transfers the knowledge to the feature extractor from the same group, acting as a bridge which connects the feature extractor to the sentiment classifier from the other group.
In particularly, our mutual learning based loss functions are defined as follows:
\begin{eqnarray}\label{eq:ml-loss}
  \loss_{ML_1} &=& \sum_{i=n^s+1}^{n^s+n^t} \mathrm{D}_{KL}\big(\Classifier_2(\dvec_i) \| \Prober_1(\dvec_i)\big), \label{eq:ml_1}\\ 
  \loss_{ML_2} &=& \sum_{i=n^s+1}^{n^s+n^t} \mathrm{D}_{KL}\big(\Classifier_1(\dvec_i) \| \Prober_2(\dvec_i)\big).\label{eq:ml_2}
\end{eqnarray}

Correspondingly, we have two new versions of objectives $\loss_{G_1}$ and $\loss_{G_2}$ by replacing $\loss_{sML_1}$ and $\loss_{sML_2}$ with $\loss_{ML_1}$ and $\loss_{ML_2}$.
Taking $\Theta_{C_1}$ for illustration, compared with standard mutual learning, our approach changes the gradients of the classifiers from $\frac{\partial \loss_{CLS_1}}{\partial \Theta_{C_1}} + \lambda_{M} \frac{\partial \loss_{sML_1}}{\partial \Theta_{C_1}}$ to $\frac{\partial \loss_{CLS_1}}{\partial \Theta_{C_1}}$.
Thus it frees the classifiers from learning not very reliable pseudo-labels as mentioned before. 

Thanks to the integration of adversarial learning and mutual learning, the gradients of feature extractors are given as: $\frac{\partial \loss_{G_1}}{\partial \Theta_{FE_1}}=\frac{\partial \loss_{CLS_1}}{\partial \Theta_{FE_1}} - \eta\lambda_{D}\frac{\partial \loss_{DOM_1}}{\partial \Theta_{FE_1}}+\lambda_{M}\frac{\partial \loss_{ML_1}}{\partial \Theta_{FE_1}}$ and $\frac{\partial \loss_{G_2}}{\partial \Theta_{FE_2}}=\frac{\partial \loss_{CLS_2}}{\partial \Theta_{FE_2}} - \eta\lambda_{D}\frac{\partial \loss_{DOM_2}}{\partial \Theta_{FE_2}}+\lambda_{M}\frac{\partial \loss_{ML_2}}{\partial \Theta_{FE_2}}$, enabling the feature extractors to learn data distributions and sentiment information from both the source domain and the target domain.

\section{Experiments}
In this section, we assess the effectiveness of our approach DAML by first clarifying the experimental setup and afterwards analyzing the experimental results.

\subsection{Experimental Setup}
\subsubsection{Datasets} We adopt multiple publicly available datasets with different orgins to evaluate DAML.
The first pair of source and target datasets is Yelp and IMDB datasets built by~\cite{tang2015learning}.
One issue is that Yelp has 5 sentiment labels, while the number is 10 for IMDB.
To align the space of sentiment labels for domain adaptation, we simply divide the scores of the sentiment labels in IMDB by 2 and round them up.

Moreover, to investigate the performance in different domains with the same origin, we choose three domains, i.e., Electronics, CD, and Clothing, from the Amazon dataset~\cite{McAuleyTSH15}.
For each domain, there are 80,000 pieces of documents in the training set and 10,000 in
both the test set and development set.
All of these statistics are summarized in Table~\ref{tab:statistics} in detail.

\begin{table}[!t]
\centering
\resizebox{.85\columnwidth}{!}{
  \begin{tabular}{|c|c|c|c|c|c|}
    \hline
    Domain & Training & Development & Test \\ \hline
    Yelp       &  62,522  & 7,773       & 8,671\\ \hline
    IMDB        &  67,426  & 8,381       & 9,112 \\ \hline
    Electronics &  79,942  & 9,994       & 9,994 \\ \hline
    CD          & 79,999   & 10,000      & 10,000 \\ \hline
    Clothing    &  79,990  & 10,000      & 10,000 \\ \hline
  \end{tabular}}
  \caption{Statistics of the experimental datasets.}  \label{tab:statistics}
\end{table}

\begin{table*}[t]
  \centering
 \caption{Sentiment classification results in terms of Acc and RMSE. 
The best results are marked in bold.}\label{tab:full-comparison}
\begin{tabular}{ll|c|c|c|c|c|c|c|c|c|c}
\hline
\multicolumn{1}{l|}{Source} &                     & \multicolumn{6}{c|}{Baselines} & \multicolumn{4}{c}{Our Approaches}                                                    \\ \cline{3-12}
  \multicolumn{1}{c|}{$\hookrightarrow$ Target} & Metrics & \multicolumn{1}{c|}{Naive} & \multicolumn{1}{c|}{DANN} & \multicolumn{1}{c|}{MMD} & \multicolumn{1}{c|}{WDGRL} & \multicolumn{1}{c|}{DAS} & \multicolumn{1}{c|}{ACAN} & \multicolumn{1}{c|}{NE} & \multicolumn{1}{c|}{ML} & \multicolumn{1}{c|}{FA} & DAML \\ \hline \hline
\multicolumn{1}{l|}{Yelp }                           & Acc  &0.513 &0.528          &0.525   &0.522          &0.471 &0.508 &0.532          &0.529 &0.519 &\textbf{0.563} \\ \cline{2-2}
  \multicolumn{1}{c|}{$\hookrightarrow$ IMDB}        & RMSE &0.956 &0.867          &0.877   &0.882          &1.029 &0.967 &0.867          &0.907 &0.925 &\textbf{0.822} \\ \cline{1-2}
  \multicolumn{1}{l|}{IMDB}                          & Acc  &0.510 &0.522          &0.511   &0.494          &0.509 &0.502 &0.518          &0.517 &0.509 &\textbf{0.528} \\ \cline{2-2}
  \multicolumn{1}{c|}{ $\hookrightarrow$ Yelp}       & RMSE &0.921 &0.881          &0.971   &0.955          &0.929 &0.964 &\textbf{0.865} &0.889 &0.883 &0.904 \\ \cline{1-2}
  \multicolumn{1}{l|}{CD }                           & Acc  &0.662 &0.664          &0.655   &0.655          &0.627 &0.657 &0.653          &0.660 &0.654 &\textbf{0.669} \\ \cline{2-2}
  \multicolumn{1}{c|}{$\hookrightarrow$ Electronics} & RMSE &1.074 &\textbf{1.016} &1.046   &1.097          &1.123 &1.109 &1.077          &1.030 &1.059 &1.068 \\ \cline{1-2}
  \multicolumn{1}{l|}{Electronics }                  & Acc  &0.654 &0.654          &0.649   &0.655          &0.628 &0.654 &0.659          &0.657 &0.655 &\textbf{0.670} \\ \cline{2-2}
  \multicolumn{1}{c|}{$\hookrightarrow$ CD}          & RMSE &1.020 &0.970          &1.021   &0.986          &1.086 &1.023 &0.943          &0.960 &1.003 &\textbf{0.922} \\ \cline{1-2}
  \multicolumn{1}{l|}{CD }                           & Acc  &0.658 &0.657          &0.651   &0.648          &0.582 &0.656 &0.660          &0.622 &0.621 &\textbf{0.667} \\ \cline{2-2}
  \multicolumn{1}{c|}{$\hookrightarrow$ Clothing}    & RMSE &\textbf{0.920} &0.975 &1.044   &1.009          &1.359 &0.974 &0.993          &1.190 &1.180 &0.953 \\ \cline{1-2}
  \multicolumn{1}{l|}{Clothing }                     & Acc  &0.644 &0.639          &0.650   &0.634          &0.596 &0.650 &0.650          &0.610 &0.610 &\textbf{0.661} \\ \cline{2-2}
  \multicolumn{1}{c|}{$\hookrightarrow$ CD }         & RMSE &1.067 &1.034          &1.021   &0.919          &1.285 &1.030 &0.957          &1.246 &1.229 &\textbf{0.917} \\ \cline{1-2}
  \multicolumn{1}{l|}{Electronics }                  & Acc  &0.692 &0.693          &0.689   &0.693          &0.582 &0.693 &0.702 &0.690 &0.672 &\textbf{0.705} \\ \cline{2-2}
  \multicolumn{1}{c|}{$\hookrightarrow$ Clothing}    & RMSE &0.819 &0.851          &0.840   &0.805          &1.359 &0.827 &0.806          &0.857 &0.929 &\textbf{0.781} \\ \cline{1-2}
  \multicolumn{1}{l|}{Clothing }                     & Acc  &0.675 &0.689          &0.680   &0.678          &0.594 &0.685 &0.696 &0.663 &0.666 &\textbf{0.698} \\ \cline{2-2}
  \multicolumn{1}{c|}{$\hookrightarrow$ Electronics} & RMSE &0.878 &0.883          &0.911   &0.913          &1.424 &0.887 &0.891          &1.086 &1.033 &\textbf{0.819} \\ \cline{1-2}
  \multicolumn{1}{c|}{Average}     & Acc  &0.626 &0.631 &0.626 &0.622 &0.547 &0.626 &0.634 &0.619 &0.613 &\textbf{0.645} \\ \cline{2-2}
  \multicolumn{1}{c|}{Performance} & RMSE &0.957 &0.935 &0.966 &0.946 &1.199 &0.972 &0.925 &1.027 &1.030 &\textbf{0.898} \\ \cline{1-2}
  \hline
\end{tabular}
\end{table*}

\subsubsection{Evaluation}
Following the study of~\cite{tang2015learning}, the model performance in our experiments is compared according to
$Acc$, short for $Accuracy$, which measures the classification performance, and $RMSE$, which
shows the divergence between the predicted ratings and the ground truth ratings. 

For unsupervised domain-adaptation task, the unavailability of the development set in target domains results in two issues.
First, it is impossible to tune hyper-parameters for every task according to the performance on the development set of each target domain.
Therefore, as is done by~\cite{das}, we build a development set available for only one task (Yelp $\rightarrow$ IMDB). 
Hyper-parameters of our model and baselines are tuned to perform their best on this development set and then fixed on all tasks.
Second, we are not able to do early-stopping for a model according to its performance in a target domain. 
As an alternative, we use the development set of each source domain to determine early-stopping.
Specifically, for each model, we save its parameters when it achieves better results on the development set along the training process. 
And finally the saved parameters with best performance are regarded as model parameters for testing on the target domain.
All the models experimented in this work determine their model parameters through the above manner.

For each mutual learning based approach, the classifier that performs the best on the development set of a source domain is chosen to be eventually evaluated on a target domain.

\subsubsection{Implementations}
As mentioned in the section of feature extraction, our approach uses HAN as the feature extractor for its good performance.
For a fair comparison, we make other approaches to use HAN as well.
Though Bidirectional Encoder Representations from Transformers (BERT)~\cite{bert} has been gaining more and more attention as a powerful text feature extractor, we did not find an appropriate setting to obtain significantly better results than HAN from our local experiments.

Before the training process starts, 200-dimensional word embeddings are learned by Word2vec~\cite{mikolov2013distributed}. Word2vec is trained on all the available data in all the experimented domains. 
The hyper-parameters are fixed for all domains. By default, $\eta$, $\lambda_D$, $\lambda_M$ are set as 0.005, 1.0 and 1.0, respectively.
Adam is adopted as the optimizer for all the experiments.

For the baselines, if their source codes are available, we directly utilize them by just adjusting input and tuning parameters.
Otherwise, we implement them by referring to the settings shown in their papers.

\subsubsection{Baselines}
The following end-to-end domain-adapted sentiment classification models are adopted for comparison.

\noindent \textbf{Naive}: We experiment with a sentiment classification model implemented by
HAN without any domain-adaptation technique.
It is trained on a source domain and then directly tested on a target domain.

\noindent \textbf{DANN} \cite{dann}: An adversarial discriminator is introduced to distinguish data from different domains so that its feature extractor could finally generate domain-invariant latent feature vectors by trying to confuse the discriminator. 

\noindent \textbf{MMD} \cite{gretton2012kernel}: As another common method for domain adaptation,
MMD is applied to a feature extractor to make it domain-independent. 
The results of MMD are mainly compared with the ones of adversarial domain adaptation and so as to examine the effectiveness 
of the latter.

\noindent \textbf{WDGRL} \cite{wdgrl}: This approach is similar to DANN, but uses Wasserstein Distance instead of JS-divergence as the loss function of the discriminator.

\noindent \textbf{DAS} \cite{das}\footnote{https://github.com/ruidan/DAS}: 
This approach employs entropy minimization
and a self-ensemble bootstrapping method to refine its classifier while minimizing the
domain divergence.

\noindent \textbf{ACAN} \cite{acan}\footnote{https://github.com/XiaoYee/ACAN}: ACAN introduces two classifier networks on the top of the feature extractor. The discrepancy of two classifiers is increased to provide diverse views, while the extractor learns to create better features away from the category boundaries by minimizing this discrepancy.

To fully investigate the mechanism behind our approach, we consider the following variants: 

\noindent \textbf{Naive Ensemble (NE)}: To show that DAML performs well in terms of leveraging multiple parallel models, we design a naive ensemble framework which fetches the probabilities returned by two domain-adapted models and takes their addition as its prediction.

\noindent \textbf{ML} \cite{dml}: It is the direct application of standard mutual learning with adversarial learning to our task, as discussed in previous section. 

\noindent \textbf{Feature Alignment (FA)}:
This method aligns two classification models at the level of latent feature layer by minimizing the Euclidean distance between the outputs of two feature extractors for each input document.

\subsection{Experimental Results}

\subsubsection{Model comparison}
Table~\ref{tab:full-comparison} compares our approaches with baselines on different domain-adaptation tasks.
Our final approach DAML outperforms all the other approaches on most tasks.
Compared with the NE framework, which in average performs the best among all the left approaches, DAML improves from 63.4\% to 64.5\% in terms of accuracy and reduces the RMSE error from 0.925 to 0.898.  
Moreover, NE requires multiple available classification models when testing, while DAML only needs a single model that performs the best on the development set of a source domain.

Among the four proposed methods, the ML and FA learning frameworks perform significantly worse than NE and DAML, and even worse than single models.
As is analyzed before, ML requires each model to trust its weak peer.
Its failure emphasizes the intuition that models
ought to take different views on ground truth and pseudo-labels.
On the other hand, though FA tries to align multiple models on the layer of the feature representation without disturbing their classifiers, it does not provide models with capability on learning from each other effectively.

\begin{table}[!t]
  \centering
  \caption{Results of DAML trained on different domains, with its probers exclusively applied to the domain indicated by the column name.}\label{tab:prober-domain}
\resizebox{\columnwidth}{!}{\begin{tabular}{ll|c|c|c}
\hline
  \multicolumn{1}{l|}{Source}                        &         & \multicolumn{1}{c|}{Target} & \multicolumn{1}{c|}{Source} & \multicolumn{1}{c}{Both}\\
  \multicolumn{1}{c|}{$\hookrightarrow$ Target}      & Metrics & \multicolumn{1}{c|}{Domain} & \multicolumn{1}{c|}{Domain} & \multicolumn{1}{c}{Domains} \\ \hline \hline
  \multicolumn{1}{l|}{Yelp }                         & Acc     & \textbf{0.563}              & 0.521                       & 0.546          \\ \cline{2-2}
  \multicolumn{1}{c|}{$\hookrightarrow$ IMDB}        & RMSE    & \textbf{0.822}              & 0.895                       & 0.827          \\ \cline{1-2}
  \multicolumn{1}{l|}{IMDB}                          & Acc     & 0.528                       & 0.514                       & \textbf{0.531} \\ \cline{2-2}
  \multicolumn{1}{c|}{ $\hookrightarrow$ Yelp}       & RMSE    & 0.904                       & 0.890                       & \textbf{0.852} \\ \cline{1-2}
  \multicolumn{1}{l|}{CD }                           & Acc     & \textbf{0.669}              & 0.652                       & 0.663          \\ \cline{2-2}
  \multicolumn{1}{c|}{$\hookrightarrow$ Electronics} & RMSE    & 1.068                       & 1.085                       & \textbf{1.015} \\ \cline{1-2}
  \multicolumn{1}{l|}{Electronics }                  & Acc     & \textbf{0.670}              & 0.669                       & 0.664          \\ \cline{2-2}
  \multicolumn{1}{c|}{$\hookrightarrow$ CD}          & RMSE    & \textbf{0.922}              & 0.942                       & 0.946          \\ \cline{1-2}
  \multicolumn{1}{c|}{Average}                       & Acc     & \textbf{0.608}              & 0.589                       & 0.601          \\ \cline{2-2}
  \multicolumn{1}{c|}{Performance}                   & RMSE    & 0.929                       & 0.953                       & \textbf{0.910} \\ \cline{1-2}
  \hline
\end{tabular}
}
\end{table}

\subsubsection{Mutual learning on different domain settings}
So far, one of the most important remaining questions is whether our learning framework works by offering pseudo-labels of the target domain or by only introducing a regularizer that pulls two models closer.
To answer this question, we try to train a group of the DAML frameworks which apply probers to only the source domain, only the target domain, and both the two domains.
These experiments are conducted on the Yelp, IMDB, Amazon Electronics, and CD datasets.

Table~\ref{tab:prober-domain} presents the corresponding results. 
We can see training the probers of DAML on the source domain exclusively performs much worse than the ones trained on the target domain, or both domains.
The comparison emphasizes the importance of leveraging the sentiment information in the target domains.
According to the results, we can conclude that what makes DAML perform well is its ability of taking an advantage of pseudo-labels.

\begin{figure}[!t]
\centering
\begin{subfigure}[c]{0.495\columnwidth}
    \includegraphics[width=\linewidth]{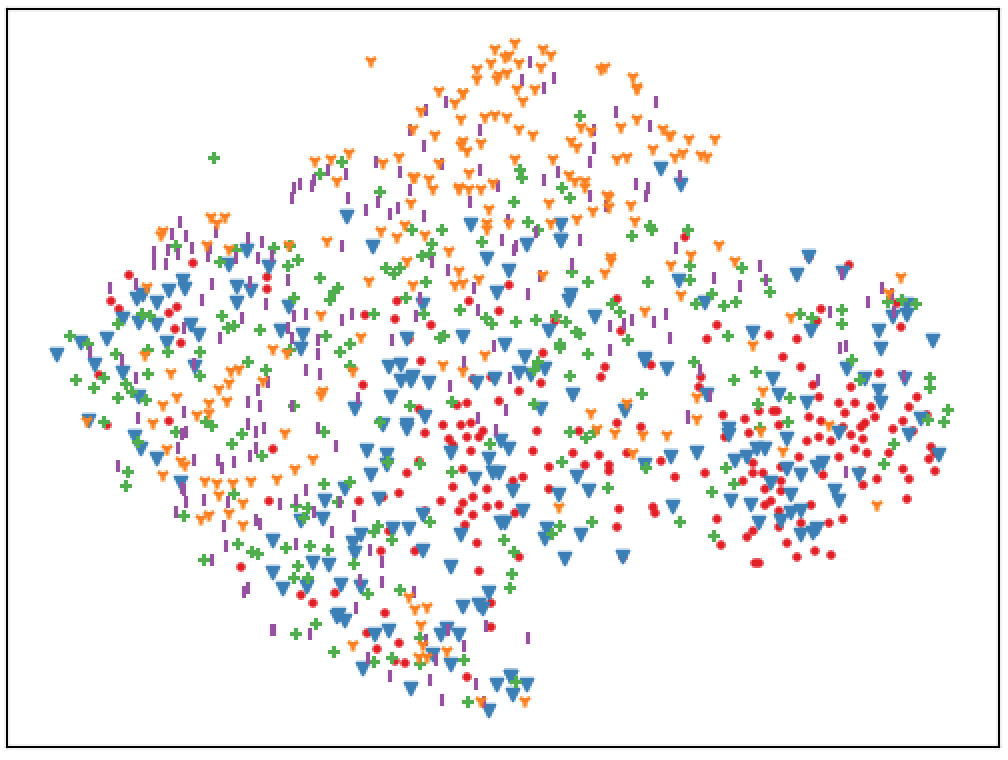}
    \caption{Target feature: Standard ML.}
    \label{subfig:vis-ml}
\end{subfigure}
\begin{subfigure}[c]{0.495\columnwidth}
    \includegraphics[width=\linewidth]{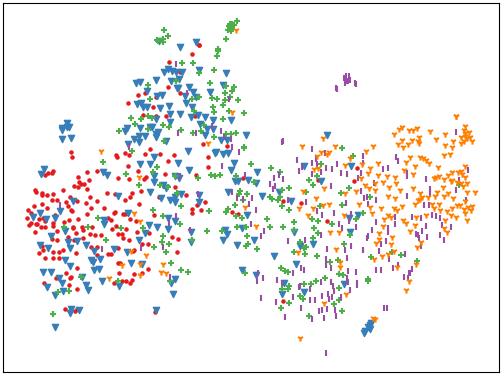}
    \caption{Target feature: DAML.}
    \label{subfig:vis-daml}
\end{subfigure}
\caption{t-SNE visualization of the features in target domain learned by standard Mutual Learning and proposed DAML methods. Different rating scores are indicated by colors and shapes. (Red = 1, Blue = 2, Green = 3, Purple = 4, Yellow = 5)}\label{fig:vis}
\end{figure}

\subsubsection{Feature visualization}
Taking the task of Yelp $\rightarrow$ IMDB for illustration, we visualize the feature vectors of the target domain generated by our approach and standard mutual learning (ML).
As shown in Figure~\ref{fig:vis}, the feature representations learned by ML have vague classification boundaries.
By contrast, our approach successfully reveals the sentiment polarity in the obtained representations, since negative (red and blue), neutral (green), and positive (purple and yellow) examples are located in the left, middle, and right parts of the figure, respectively.

\subsubsection{Analysis on training process}
\begin{figure}[!t]
\begin{subfigure}[c]{0.48\columnwidth}
    \includegraphics[width=\linewidth]{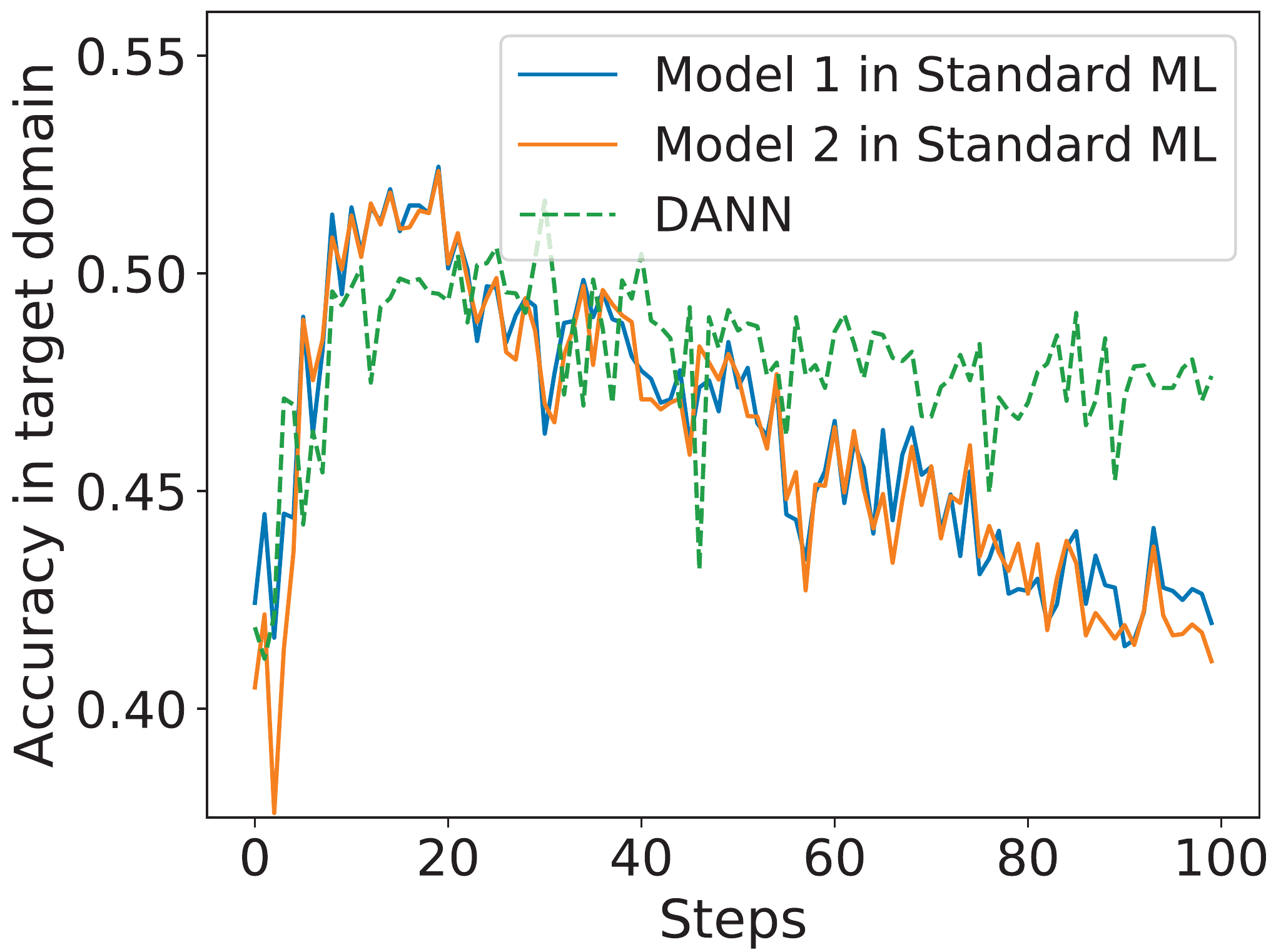}
    \caption{Standard ML vs. DANN.}
    \label{fig:ml-tr-acc}
\end{subfigure}
\begin{subfigure}[c]{0.48\columnwidth}
    \includegraphics[width=\linewidth]{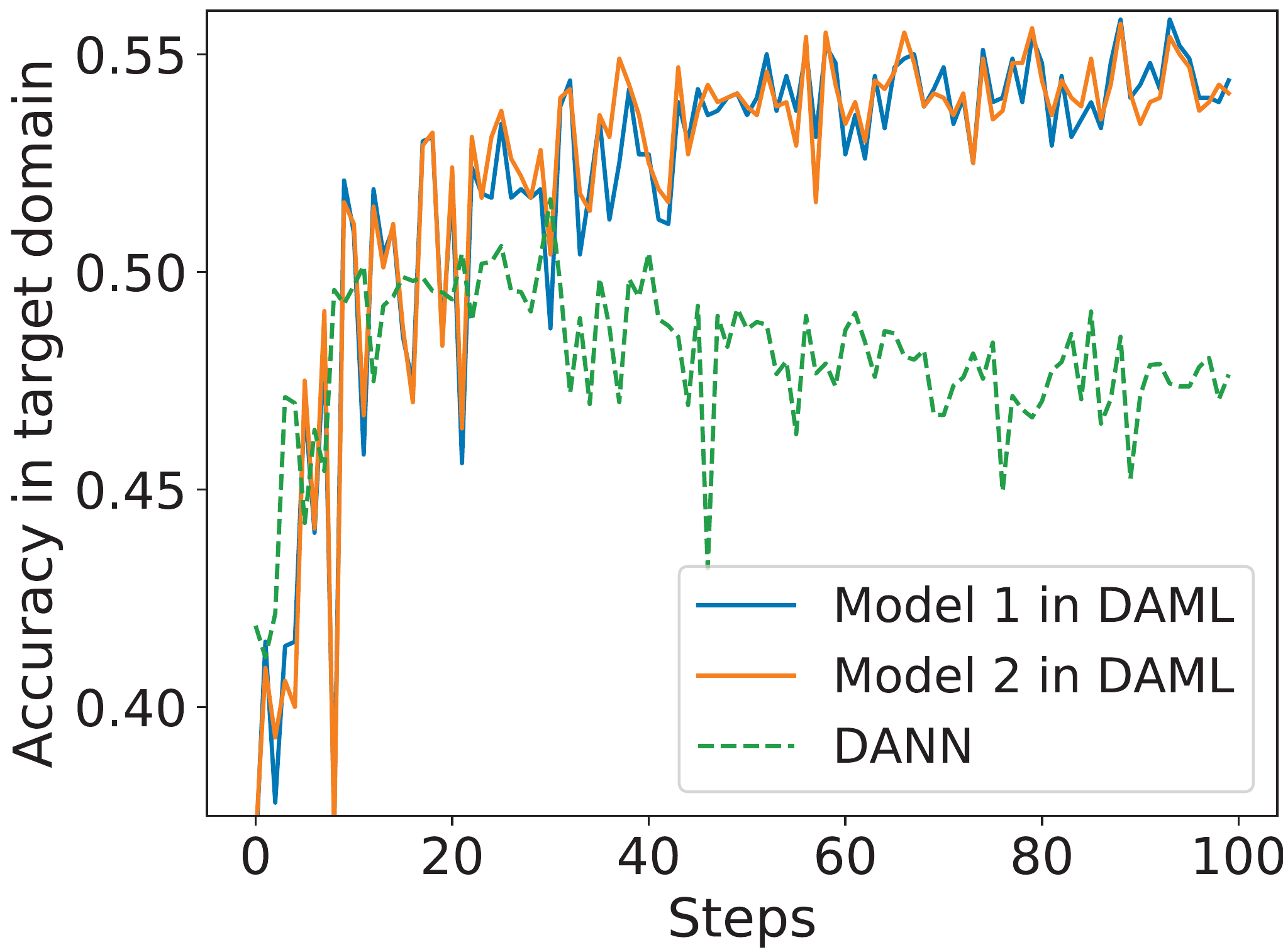}
    \caption{DAML vs. DANN.}
    \label{fig:mp-tr-acc}
\end{subfigure}
\caption{Accuracy on the test set, evaluated at each training step. 
In each picture, two solid lines stand for the two models trained in the framework.
The dashed line represents the plain adversarial domain-adaptation method DANN.}
\label{fig:test-acc}
\end{figure}

It can help understand the effect of each mutual learning framework to see its performance in the target domain at each step. Therefore, we conduct a series of experiments on the Yelp$\rightarrow$IMDB task, where a development set from the target domain is available. The performance on the development set of the chosen approaches is evaluated every several steps and the curves of accuracy are drawn in Figure~\ref{fig:test-acc}.

We can see DANN and two sentiment classifiers in the two mutual learning based frameworks obtain high performance at about the twentieth steps. 
After that, DANN keeps its performance relatively stable, compared with two other models in the figure.
Particularly, the curves of standard ML start to drop rapidly, while our final approach gradually
obtains even higher performance.
These comparisons clearly illustrate the negative influence of trusting each other's weak domain-adapted classifier to much.

\subsubsection{Influence of hyper-parameter $\eta$ and $\lambda_M$}
In order to observe the influence of the relative weights of different parts in the total loss function, we conduct some experiments with the development
sets of the target domains. Specifically, three tasks, i.e., Yelp $\rightarrow$
IMDB, Clothing $\rightarrow$ Electronics and Electronics $\rightarrow$ CD, are chosen.
For each task, our learning approach is trained and then validated on the development set of the target domain for just showing the influence. 
This procedure is repeated several times with different $\eta$ and $\lambda_M$.
$\lambda_D$ is not considered because it only influences the performance of the discriminators when $\eta$ controls the gradient flowed to the feature extractors.

\begin{figure}[!t]
\begin{subfigure}[c]{0.48\columnwidth}
    \includegraphics[width=\linewidth]{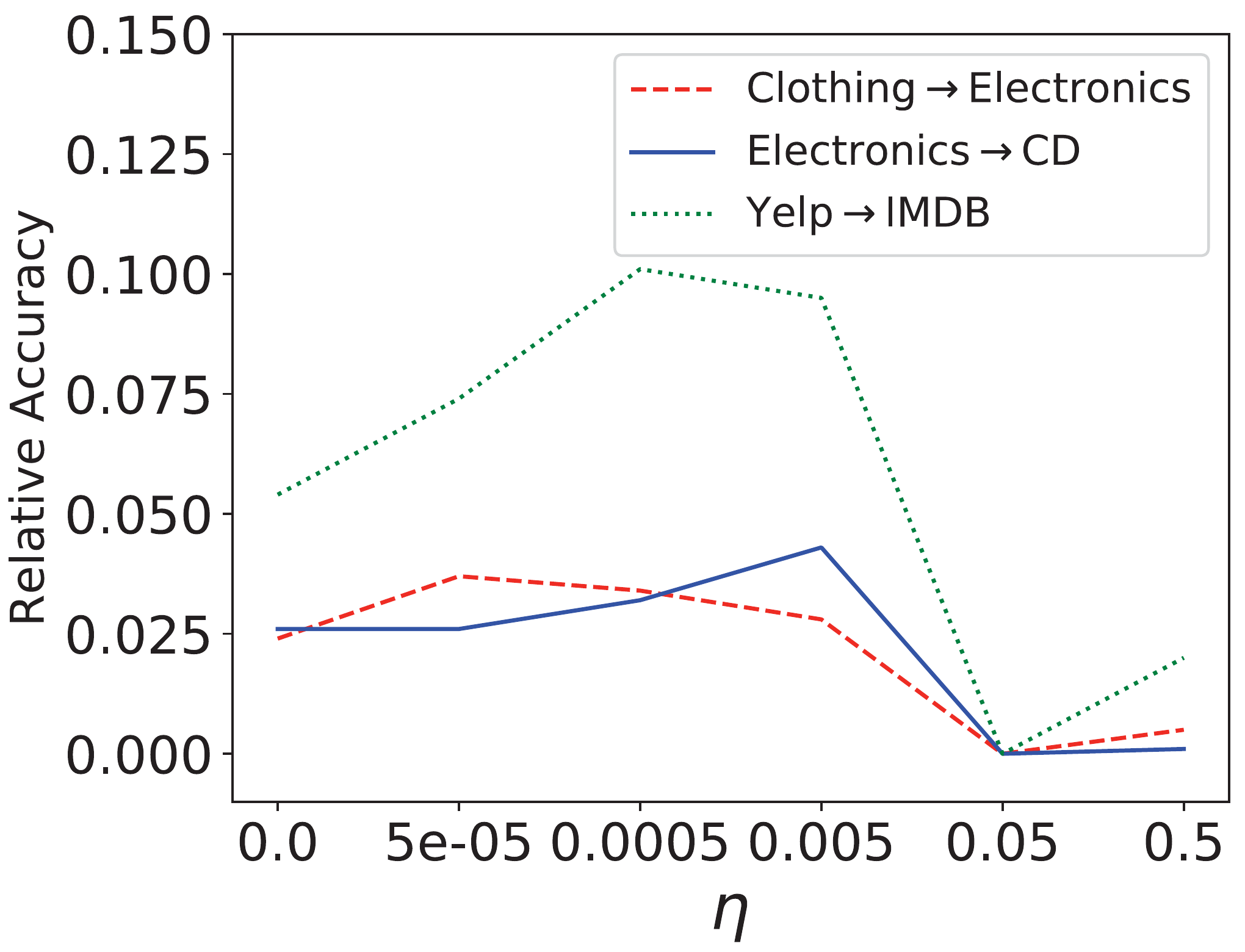}
    \caption{Influence of $\eta$.}
    \label{subfig:eta}
\end{subfigure}
\begin{subfigure}[c]{0.48\columnwidth}
    \includegraphics[width=\linewidth]{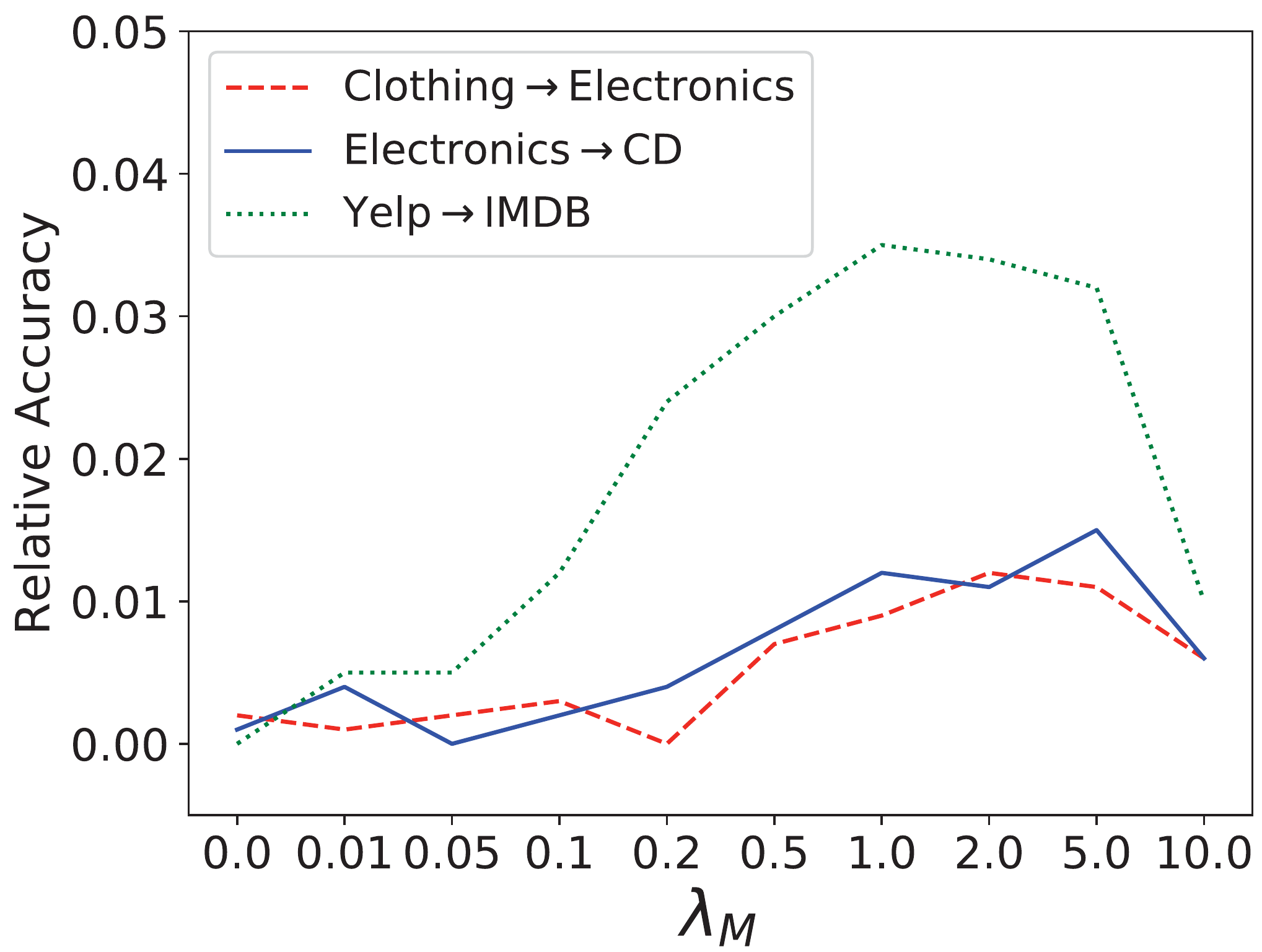}
    \caption{Influence of $\lambda$.}
    \label{subfig:lambda}
\end{subfigure}
\caption{DAML trained on three tasks with different $\eta$ and $\lambda_M$. The vertical axis stands for the relative accuracy value gained w.r.t. the minimum point for each curve.}
\label{fig:training-acc}
\end{figure}

Figure~\ref{fig:training-acc} depicts how the change of $\eta$ and $\lambda_M$ influences the performance.
It can be seen from the left subfigure that DAML does not perform very well when adversarial learning is not applied (when $\eta$ is 0), which shows the contribution of adversarial learning to our learning framework.
Moreover, we can observe from the right subfigure that DAML obtains good performance when $\lambda_M$ is in the range $[1.0, 5.0]$ for all the three tasks.
Besides, compared with the standard domain-adaptation model without mutual learning  (when $\lambda_M$ equals to 0), DAML gains improvement when $\lambda_M$ is in even a larger range.
These curves show the effectiveness of our learning framework and its robustness in terms of the critical hyper-parameters.

\subsubsection{Influence of number of groups and models}
In the end, we concisely analyze how the results change by increasing the number of groups in DAML and domain-adapted models in the baseline NE, from 2 to 4.
We take the domains of Yelp and IMDB for testing and show the performance in 
Table~\ref{tab:more-groups}.

\begin{table}[!h]
  \centering
 \caption{Sentiment classification results of more groups and models. }\label{tab:more-groups}
\resizebox{\columnwidth}{!}{\begin{tabular}{ll|c|c|c|c|c|c}
\hline
\multicolumn{1}{l|}{Source} & & \multicolumn{3}{c|}{DAML} & \multicolumn{3}{c}{NE}                                                    \\ \cline{3-8}
  \multicolumn{1}{c|}{$\hookrightarrow$ Target} & Metrics & \multicolumn{1}{c|}{2} & \multicolumn{1}{c|}{3} & \multicolumn{1}{c|}{4} & \multicolumn{1}{c|}{2} & \multicolumn{1}{c|}{3} & \multicolumn{1}{c}{4} \\ \hline \hline
\multicolumn{1}{l|}{Yelp }                           & Acc  &\textbf{0.563} &\textbf{0.563}          &0.547   &0.532          &0.527 &0.534  \\ \cline{2-2}
  \multicolumn{1}{c|}{$\hookrightarrow$ IMDB}        & RMSE &0.822 &\textbf{0.819}          &0.840   &0.867          &0.874 &0.887  \\ \cline{1-2}
  \multicolumn{1}{l|}{IMDB}                          & Acc  &0.528 &\textbf{0.531}          &0.522   &0.518          &0.515 &0.506  \\ \cline{2-2}
  \multicolumn{1}{c|}{ $\hookrightarrow$ Yelp}       & RMSE &0.904 &0.874          &0.916   &\textbf{0.865}          &0.903 &0.876 \\
  \hline
\end{tabular}}
\end{table}

It can be seen that, in this work, involving more models doesn't always improve the performance of the approaches.
This might be attributed to the fact that a development set from a target domain is hardly available in practice, as mentioned in the section of evaluation. 
Therefore, we follow previous work to train each model until it performs the best on a source domain.
In consequence, though an approach with more groups or models may perform better on the target domain at some training points during the learning process, we are not able to finally obtain these peek points.

\section{Conclusion}
In this work, we have devised DAML, a novel deep adversarial mutual learning framework through which two domain-adapted models can learn from each other to effectively utilize the sentiment information lying in a target domain.
This advantage is enjoyed by the means of incorporating probers into mutual learning, which act as bridges to connect feature extractors with sentiment classifiers.
Unlike standard mutual learning, our learning framework introduces a flexible manner for each model to decide how to view its peer's predictions, rather than to align the output labels of sentiment classifiers, which is somewhat questionable in unsupervised domain-adaptation tasks. 
Experiments have shown that DAML significantly improves the performance over representative end-to-end models for domain-adapted sentiment classification.

\section{Acknowledgments}
We thank the anonymous reviewers for their valuable comments.
This work is supported by National Key Research and Development Program (2019YFB2102600), NSFC (61702190), NSFC-Zhejiang (U1609220), and Shanghai Sailing Program (17YF1404500).

\bibliographystyle{aaai}
\small
\bibliography{AAAI-XueQ.1846}

\begin{thebibliography}{}

\bibitem[\protect\citeauthoryear{Bahdanau, Cho, and
  Bengio}{2015}]{BahdanauCB14}
Bahdanau, D.; Cho, K.; and Bengio, Y.
\newblock 2015.
\newblock Neural machine translation by jointly learning to align and
  translate.
\newblock In {\em ICLR}.

\bibitem[\protect\citeauthoryear{Ben{-}David \bgroup et al\mbox.\egroup
  }{2010}]{Ben-DavidBCKPV10}
Ben{-}David, S.; Blitzer, J.; Crammer, K.; Kulesza, A.; Pereira, F.; and
  Vaughan, J.~W.
\newblock 2010.
\newblock A theory of learning from different domains.
\newblock {\em Machine Learning} 79(1-2):151--175.

\bibitem[\protect\citeauthoryear{Blitzer, Dredze, and
  Pereira}{2007}]{BlitzerDP07}
Blitzer, J.; Dredze, M.; and Pereira, F.
\newblock 2007.
\newblock Biographies, bollywood, boom-boxes and blenders: Domain adaptation
  for sentiment classification.
\newblock In {\em ACL}.

\bibitem[\protect\citeauthoryear{Chen \bgroup et al\mbox.\egroup
  }{2016}]{Chen-EMNLP2016}
Chen, H.; Sun, M.; Tu, C.; Lin, Y.; and Liu, Z.
\newblock 2016.
\newblock Neural sentiment classification with user and product attention.
\newblock In {\em EMNLP},  1650--1659.

\bibitem[\protect\citeauthoryear{Devlin \bgroup et al\mbox.\egroup
  }{2019}]{bert}
Devlin, J.; Chang, M.; Lee, K.; and Toutanova, K.
\newblock 2019.
\newblock {BERT:} pre-training of deep bidirectional transformers for language
  understanding.
\newblock In {\em NAACL},  4171--4186.

\bibitem[\protect\citeauthoryear{Ganin and Lempitsky}{2015}]{GaninL15}
Ganin, Y., and Lempitsky, V.~S.
\newblock 2015.
\newblock Unsupervised domain adaptation by backpropagation.
\newblock In {\em ICML},  1180--1189.

\bibitem[\protect\citeauthoryear{Ganin \bgroup et al\mbox.\egroup
  }{2016}]{dann}
Ganin, Y.; Ustinova, E.; Ajakan, H.; Germain, P.; Larochelle, H.; Laviolette,
  F.; Marchand, M.; and Lempitsky, V.
\newblock 2016.
\newblock Domain-adversarial training of neural networks.
\newblock {\em JMLR} 17(1):2096--2030.

\bibitem[\protect\citeauthoryear{Glorot, Bordes, and
  Bengio}{2011}]{glorot2011domain}
Glorot, X.; Bordes, A.; and Bengio, Y.
\newblock 2011.
\newblock Domain adaptation for large-scale sentiment classification: A deep
  learning approach.
\newblock In {\em ICML},  513--520.

\bibitem[\protect\citeauthoryear{Gretton \bgroup et al\mbox.\egroup
  }{2012a}]{gretton2012kernel}
Gretton, A.; Borgwardt, K.~M.; Rasch, M.~J.; Sch{\"o}lkopf, B.; and Smola, A.
\newblock 2012a.
\newblock A kernel two-sample test.
\newblock {\em JMLR} 13(Mar):723--773.

\bibitem[\protect\citeauthoryear{Gretton \bgroup et al\mbox.\egroup
  }{2012b}]{GrettonSSSBPF12}
Gretton, A.; Sriperumbudur, B.~K.; Sejdinovic, D.; Strathmann, H.;
  Balakrishnan, S.; Pontil, M.; and Fukumizu, K.
\newblock 2012b.
\newblock Optimal kernel choice for large-scale two-sample tests.
\newblock In {\em NIPS},  1214--1222.

\bibitem[\protect\citeauthoryear{He \bgroup et al\mbox.\egroup }{2018}]{das}
He, R.; Lee, W.~S.; Ng, H.~T.; and Dahlmeier, D.
\newblock 2018.
\newblock Adaptive semi-supervised learning for cross-domain sentiment
  classification.
\newblock In {\em EMNLP},  3467--3476.

\bibitem[\protect\citeauthoryear{Hinton, Vinyals, and
  Dean}{2015}]{hinton2015distilling}
Hinton, G.; Vinyals, O.; and Dean, J.
\newblock 2015.
\newblock Distilling the knowledge in a neural network.
\newblock {\em arXiv preprint arXiv:1503.02531}.

\bibitem[\protect\citeauthoryear{Huang \bgroup et al\mbox.\egroup
  }{2007}]{huang2007correcting}
Huang, J.; Gretton, A.; Borgwardt, K.; Sch{\"o}lkopf, B.; and Smola, A.~J.
\newblock 2007.
\newblock Correcting sample selection bias by unlabeled data.
\newblock In {\em NIPS},  601--608.

\bibitem[\protect\citeauthoryear{Kanaci \bgroup et al\mbox.\egroup
  }{2019}]{kanaci2019multi}
Kanaci, A.; Li, M.; Gong, S.; and Rajamanoharan, G.
\newblock 2019.
\newblock Multi-task mutual learning for vehicle re-identification.
\newblock In {\em CVPR Workshops},  62--70.

\bibitem[\protect\citeauthoryear{Kim}{2014}]{Kim14}
Kim, Y.
\newblock 2014.
\newblock Convolutional neural networks for sentence classification.
\newblock In {\em EMNLP},  1746--1751.

\bibitem[\protect\citeauthoryear{Liu}{2012}]{2012Liu-Book}
Liu, B.
\newblock 2012.
\newblock {\em Sentiment Analysis and Opinion Mining}.
\newblock Synthesis Lectures on Human Language Technologies. Morgan {\&}
  Claypool Publishers.

\bibitem[\protect\citeauthoryear{McAuley \bgroup et al\mbox.\egroup
  }{2015}]{McAuleyTSH15}
McAuley, J.~J.; Targett, C.; Shi, Q.; and van~den Hengel, A.
\newblock 2015.
\newblock Image-based recommendations on styles and substitutes.
\newblock In {\em SIGIR},  43--52.

\bibitem[\protect\citeauthoryear{Mikolov \bgroup et al\mbox.\egroup
  }{2013}]{mikolov2013distributed}
Mikolov, T.; Sutskever, I.; Chen, K.; Corrado, G.~S.; and Dean, J.
\newblock 2013.
\newblock Distributed representations of words and phrases and their
  compositionality.
\newblock In {\em NIPS},  3111--3119.

\bibitem[\protect\citeauthoryear{Pan and Yang}{2009}]{pan2009survey}
Pan, S.~J., and Yang, Q.
\newblock 2009.
\newblock A survey on transfer learning.
\newblock {\em IEEE TKDE} 22(10):1345--1359.

\bibitem[\protect\citeauthoryear{Peng \bgroup et al\mbox.\egroup
  }{2018}]{ZhangHPJ18}
Peng, M.; Zhang, Q.; Jiang, Y.; and Huang, X.
\newblock 2018.
\newblock Cross-domain sentiment classification with target domain specific
  information.
\newblock In {\em ACL},  2505--2513.

\bibitem[\protect\citeauthoryear{Qu \bgroup et al\mbox.\egroup }{2019}]{acan}
Qu, X.; Zou, Z.; Cheng, Y.; Yang, Y.; and Zhou, P.
\newblock 2019.
\newblock Adversarial category alignment network for cross-domain sentiment
  classification.
\newblock In {\em ACL},  2496--2508.

\bibitem[\protect\citeauthoryear{Shen \bgroup et al\mbox.\egroup
  }{2017}]{wdgrl}
Shen, J.; Qu, Y.; Zhang, W.; and Yu, Y.
\newblock 2017.
\newblock Wasserstein distance guided representation learning for domain
  adaptation.
\newblock {\em arXiv preprint arXiv:1707.01217}.

\bibitem[\protect\citeauthoryear{Socher \bgroup et al\mbox.\egroup
  }{2013}]{SocherPWCMNP13}
Socher, R.; Perelygin, A.; Wu, J.; Chuang, J.; Manning, C.~D.; Ng, A.~Y.; and
  Potts, C.
\newblock 2013.
\newblock Recursive deep models for semantic compositionality over a sentiment
  treebank.
\newblock In {\em ACL},  1631--1642.

\bibitem[\protect\citeauthoryear{Tang, Qin, and Liu}{2015a}]{Tang-emnlp15}
Tang, D.; Qin, B.; and Liu, T.
\newblock 2015a.
\newblock Document modeling with gated recurrent neural network for sentiment
  classification.
\newblock In {\em EMNLP},  1422--1432.

\bibitem[\protect\citeauthoryear{Tang, Qin, and Liu}{2015b}]{tang2015learning}
Tang, D.; Qin, B.; and Liu, T.
\newblock 2015b.
\newblock Learning semantic representations of users and products for document
  level sentiment classification.
\newblock In {\em ACL}, volume~1,  1014--1023.

\bibitem[\protect\citeauthoryear{Vincent \bgroup et al\mbox.\egroup
  }{2008}]{VincentLBM08}
Vincent, P.; Larochelle, H.; Bengio, Y.; and Manzagol, P.
\newblock 2008.
\newblock Extracting and composing robust features with denoising autoencoders.
\newblock In {\em ICML},  1096--1103.

\bibitem[\protect\citeauthoryear{Wu \bgroup et al\mbox.\egroup
  }{2018}]{Wu-AAAI18}
Wu, Z.; Dai, X.; Yin, C.; Huang, S.; and Chen, J.
\newblock 2018.
\newblock Improving review representations with user attention and product
  attention for sentiment classification.
\newblock In {\em AAAI},  5989--5996.

\bibitem[\protect\citeauthoryear{Wu \bgroup et al\mbox.\egroup
  }{2019}]{wu2019mutual}
Wu, R.; Feng, M.; Guan, W.; Wang, D.; Lu, H.; and Ding, E.
\newblock 2019.
\newblock A mutual learning method for salient object detection with
  intertwined multi-supervision.
\newblock In {\em CVPR},  8150--8159.

\bibitem[\protect\citeauthoryear{Yang \bgroup et al\mbox.\egroup }{2016}]{hrnn}
Yang, Z.; Yang, D.; Dyer, C.; He, X.; Smola, A.; and Hovy, E.
\newblock 2016.
\newblock Hierarchical attention networks for document classification.
\newblock In {\em NAACL},  1480--1489.

\bibitem[\protect\citeauthoryear{Yu and Jiang}{2016}]{YuJ16}
Yu, J., and Jiang, J.
\newblock 2016.
\newblock Learning sentence embeddings with auxiliary tasks for cross-domain
  sentiment classification.
\newblock In {\em EMNLP},  236--246.

\bibitem[\protect\citeauthoryear{Zhang and Wang}{2015}]{ZhangW15}
Zhang, W., and Wang, J.
\newblock 2015.
\newblock Prior-based dual additive latent dirichlet allocation for user-item
  connected documents.
\newblock In {\em IJCAI},  1405--1411.

\bibitem[\protect\citeauthoryear{Zhang \bgroup et al\mbox.\egroup }{2018}]{dml}
Zhang, Y.; Xiang, T.; Hospedales, T.~M.; and Lu, H.
\newblock 2018.
\newblock Deep mutual learning.
\newblock In {\em CVPR},  4320--4328.

\bibitem[\protect\citeauthoryear{Ziser and Reichart}{2018}]{Ziser-naacl28}
Ziser, Y., and Reichart, R.
\newblock 2018.
\newblock Pivot based language modeling for improved neural domain adaptation.
\newblock In {\em NAACL},  1241--1251.

\bibitem[\protect\citeauthoryear{Ziser and Reichart}{2019}]{Ziser-acl19}
Ziser, Y., and Reichart, R.
\newblock 2019.
\newblock Task refinement learning for improved accuracy and stability of
  unsupervised domain adaptation.
\newblock In {\em ACL},  5895--5906.

\end{thebibliography}

\end{document}